\documentclass[letterpaper, 10 pt, conference]{ieeeconf}  %

\IEEEoverridecommandlockouts                              %

\usepackage{graphicx} %
\usepackage{subcaption} %
\usepackage[separate-uncertainty=true,multi-part-units=single]{siunitx} %
\DeclareSIUnit\decade{dec}
\usepackage{amsmath} %
\usepackage{color}
\usepackage{colortbl}
\usepackage{soul} %
\usepackage{xcolor}

\newcommand{\highlight}[1]{#1}
\renewcommand{\hl}[1]{#1}

\newcommand\MYhyperrefoptions{bookmarks=true,bookmarksnumbered=true,%
pdfpagemode={UseOutlines},plainpages=false,pdfpagelabels=true,%
colorlinks=true,citecolor={black},%
pdftitle={Towards Low-Latency High-Bandwidth Control of Quadrotors using Event Cameras},%
pdfsubject={Robotics, Neuromorphic Engineering, Computer Vision},%
pdfauthor={R. Sugimoto, M. Gehrig, D. Brescianini, D. Scaramuzza},%
pdfkeywords={Event Cameras, Low Latency, Control}}%
\usepackage[\MYhyperrefoptions,pdftex]{hyperref}

\definecolor{somegray}{rgb}{0.5, 0.5, 0.5}
\newcommand{\darkgrayed}[1]{\textcolor{somegray}{#1}}
\makeatletter
\newcommand*\titleheader[1]{\gdef\@titleheader{#1}}
\AtBeginDocument{%
  \let\st@red@title\@title
  \def\@title{%
    \vskip-3em
    \bgroup\normalfont\large\centering\@titleheader\par\egroup
    \vskip1.5em\st@red@title}
}
\makeatother

\titleheader{\darkgrayed{This paper has been accepted for publication at the\\
IEEE International Conference on Robotics and Automation (ICRA), Paris, 2020.
\copyright IEEE}}

\title{\LARGE \bf
Towards Low-Latency High-Bandwidth Control of Quadrotors using Event Cameras
}

\author{Rika Sugimoto Dimitrova, Mathias Gehrig, Dario Brescianini, and Davide Scaramuzza%
\thanks{The authors are with the Robotics and Perception Group, Dep. of Informatics, University of Zurich, and Dep. of Neuroinformatics, University of Zurich and ETH Zurich, Switzerland--- \url{http://rpg.ifi.uzh.ch.}
This work was supported by the SNSF-ERC Starting Grant and the Swiss National Science Foundation through the National Center of Competence in Research (NCCR) Robotics.
}%
}

\begin{document}

\maketitle
\thispagestyle{empty}
\pagestyle{empty}

\begin{abstract}
Event cameras are a promising candidate to enable high speed vision-based control due to their low sensor latency and high temporal resolution.
However, purely event-based feedback has yet to be used in the control of drones.
In this work, a first step towards implementing low-latency high-bandwidth control of quadrotors using event cameras is taken. In particular, this paper addresses the problem of one-dimensional attitude tracking using a dualcopter platform equipped with an event camera.
The event-based state estimation consists of a modified Hough transform algorithm combined with a Kalman filter that outputs the roll angle and angular velocity of the dualcopter relative to a horizon marked by a black-and-white disk.
The estimated state is processed by a proportional-derivative attitude control law that computes the rotor thrusts required to track the desired attitude.
The proposed attitude tracking scheme shows promising results of event-camera-driven closed loop control:
the state estimator performs with an update rate of 1 kHz and a latency determined to be 12 ms, enabling attitude tracking at speeds of over 1600$^\circ$/s.
\end{abstract}

\section*{Supplementary material}
A video showing the closed-loop performance of the system is available at \url{https://youtu.be/3nIznSMCMtc}

\begin{figure}[t]
    \centering
    \begin{subfigure}[t]{\columnwidth}
        \centering
        \includegraphics[width=1\textwidth]{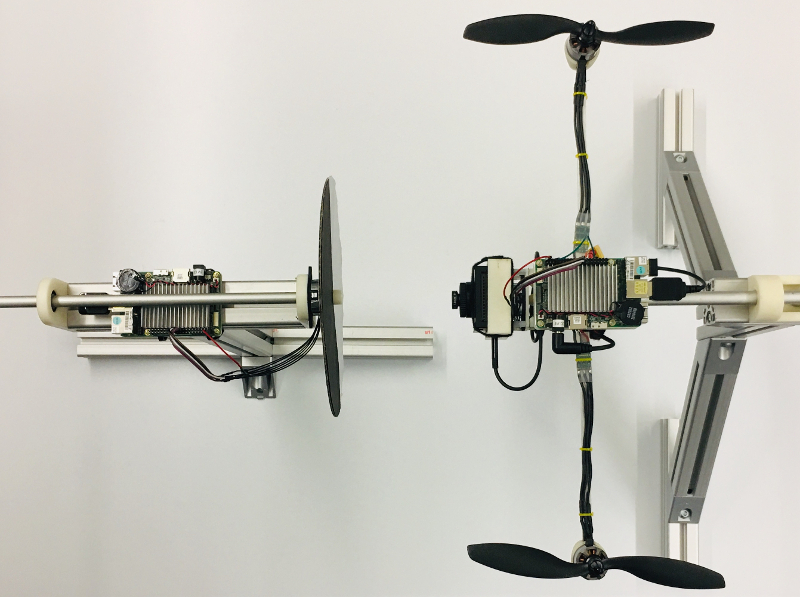}
        \caption{Top view}
    \end{subfigure}
    \begin{subfigure}[t]{0.5\columnwidth}
        \centering
        \includegraphics[width=1\textwidth]{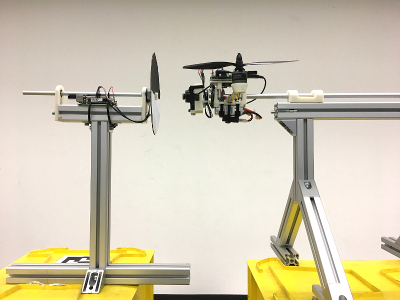}
        \caption{Side view}
    \end{subfigure}%
    \begin{subfigure}[t]{0.5\columnwidth}
        \centering
        \includegraphics[width=1\textwidth]{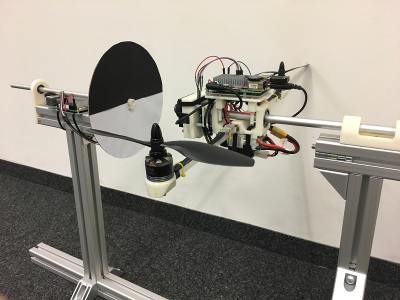}
        \caption{Perspective view}
    \end{subfigure}
    \caption{Experimental platform. The setup consists of a dualcopter equipped with an event camera and a black-and-white disk that marks the reference horizon to be tracked.}
    \label{img:platform}
\end{figure}

\section{Introduction}\label{sec:introduction}

High speed vision-based control remains a challenge in the field of robotics.
The challenge is especially pronounced on mobile platforms such as drones, which have limited power availability.
Fast flight and aggressive maneuvers require low perception latency and high controller bandwidth \cite{falanga2019}.
Traditional frame-based cameras fail to provide this low perception latency due to their limited frame rate.
They also suffer from motion blur, which further limits the speed of movements they can capture.
The recent development of low power event cameras may be key to enabling low latency vision-driven control of mobile platforms.

Event cameras are bio-inspired vision sensors that output data asynchronously \cite{gallego2019}. 
The output from these vision sensors, known as events, encode per-pixel brightness changes in the image plane.
The event camera measures such intensity changes with a microsecond resolution, making them suitable for high-speed applications where traditional cameras would fail.
Event cameras also help reduce the incoming perception information from entire images with tens of thousands of pixels to a few hundred events, making the perception pipeline more computationally efficient.

Neuromorphic-vision-driven control attempts to exploit event cameras for low latency perception and state estimation \cite{gallego2019}.
However, there is a number of challenges involved with neuromorphic-vision-driven control, particularly in the event-based state estimation step.
Events received have to be processed to make sense of what is in view, yet classic image processing is typically applied to frames and cannot be applied directly to the event stream.
One pre-processing technique is to group a number of events together to form an image of accumulated events.
Choosing the optimal time interval or number of events to accumulate is critical as it involves a trade-off between accuracy and speed:
the more events used, the more accurate the prediction is, but this comes with the cost of increased latency due to computational time and motion blur.

Another challenge with event cameras is the reduced data in stationary scenes.
When there is little change in the scene, the event camera generates fewer events, making state estimation harder and sometimes not possible.
Noise in the event stream adds to the challenge as well.

With these challenges in mind, we designed our event-based state estimator to achieve a high update rate while maintaining accuracy for both high and low speed applications.
\section{Related work}\label{sec:relatedwork} 
Event cameras have been used in state estimation in a number of simple control problems where low perception latency is critical.
An impressive application of neuromorphic-vision-driven control is the pencil-balancing robot \cite{conradt2009}.\footnote{\url{https://youtu.be/QxJ-RTbpNXw?t=215}}
The robot consisted of two event cameras and a table with arms that moved a platform in a two-dimensional plane.
The robot used an event-based adaptation of the Hough transform line detection algorithm to process the incoming events, and estimated the three-dimensional pose of the pencil with a mean update rate of over \SI{1}{\kilo\hertz}.
A hand-tuned proportional-derivative (PD) controller was used to keep the pencil upright.
Another related study is the robot goalie project, which tracked incoming balls and the position of the goalie arm with an event camera \cite{delbruck2013}.\footnote{\url{https://youtu.be/QxJ-RTbpNXw?t=229}}
Using a simple proportional controller, they were able to achieve a median update rate of \SI{550}{\hertz} and a latency of \SI{2.2}{\milli\second}.
However, these works assumed that the sensor was fixed and, therefore, the events were caused by the object to control moving in the camera field of view.
By contrast, in this work we are interested in using the event camera for feedback control where the events are mainly caused by the robot ego-motion. 

Preliminary work in closed loop control with feedback from a moving event camera was shown in \cite{mueller2015}, where, for simplification, the authors considered the one-dimensional case of controlling the yaw angle of a rotational platform.
They designed a controller that could stabilize the yaw using an efficient control update on an event-by-event fashion.
Despite being limited to 1 degree-of-freedom (DOF), the results showed that using an event camera gives improved performance, when compared to a standard camera, with respect to computational load, data rate, bandwidth, and latency.

Purely event-based perception and state estimation has yet to be used for closed loop control of drones.
A closely related work is the 6 DOF pose estimation of a hovering quadcopter \cite{mueggler2014}.
The quadcopter used an event camera to observe a stationary black square on the wall and calculate its pose relative to the square. 
The drone used an event-based Hough transform line detection algorithm over the events being tracked, and estimated its pose by minimizing the reprojection error.
The system was able to estimate its pose even under angular speeds of up to \SI{1200}{\degree\per\second}. However, the estimate was not used for closed loop control. Rather, the drone was controlled via a motion capture system.
\section{Experimental Platform}\label{sec:platform}

The objective of this work is to solve the one-dimensional attitude tracking problem of a dualcopter platform.
Fig.~\ref{img:platform} shows the experimental setup.
A black-and-white disk serves as the reference pattern, with the line between the black and white halves of the disk marking the horizon to be tracked.
The disk can be rotated by hand, and the dualcopter is to track the horizon using an event camera.

\subsection{Hardware components}
The dualcopter is equipped with DAVIS 240C, an event camera with a spatial resolution of $\text{240}\times\text{180}$ pixels and a temporal resolution of \SI{1}{\micro\second} \cite{davis}.
The camera further outputs gray-scale image frames and has a built-in 6-axis Inertial Measurement Unit (IMU); however, only the event output was used in this work for state estimation.

The DAVIS sends the event stream to an UP board\footnote{https://up-board.org/up/specifications/}, an embedded computer on the dualcopter, via USB 2.0.
The round-trip delay of the vision pipeline, from when an intensity change is triggered in the world frame to when it is detected by the embedded computer, is less than \SI{5}{\milli\second} \cite{usbdelay}. 

\begin{figure}[t]
    \centering
    \includegraphics[width=0.48\textwidth]{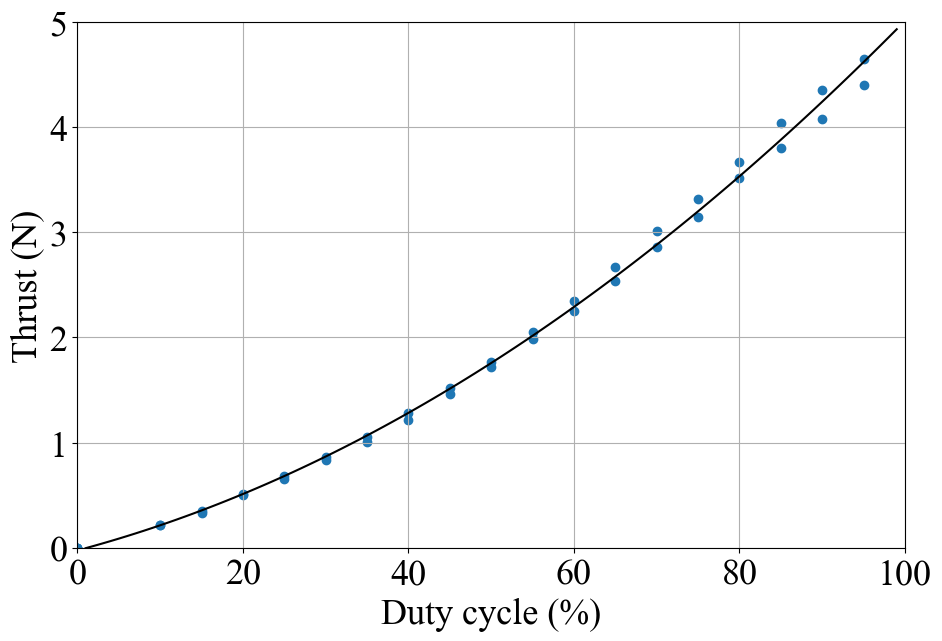}
    \caption{Thrust generated by a rotor versus the duty cycle of the pulse-width modulation (PWM) signal from the electronic speed control (ESC). We empirically determined the mapping between duty cycle and thrust generated by each rotor using a load cell. The duty cycle is proportional to the voltage delivered to the motor, thus to motor speed, and is related to thrust by a quadratic relationship. Therefore, we fit a second-degree polynomial curve through the data points.}
   \label{img:rotorthrust}
\end{figure}

The UP board processes the events for state estimation and runs the control algorithm to determine the rotor thrusts required to achieve the desired roll angle.
The controller converts the desired thrusts to the input voltage required by the motor using the thrust-to-duty-cycle mapping shown in Fig.~\ref{img:rotorthrust}.

The UP board sends motor commands to a Lumenier F4 AIO flight controller using the universal asynchronous receiver/transmitter communication protocol (UART) with a maximum transmission rate of \SI{4.1}{\kilo\hertz}.
We developed a custom firmware for the flight controller that enables communication with the motors' electronic speed control (ESC) via the DShot150 protocol at a maximum rate of \SI{9.4}{\kilo\hertz}.
The ESC sends pulse-width modulation (PWM) signals to the motors, where the input voltage to the motors is given by the applied voltage (\SI{12.1}{\volt}) times the duty cycle of the PWM signal.

The dualcopter and disk are both equipped with CUI AMT22 Modular Absolute Encoders\footnote{https://www.cui.com/product/resource/amt22.pdf} that measure the ground truth roll angle.
The encoders have an accuracy of \SI{0.2}{\degree} and a precision of \SI{0.1}{\degree}.
The encoder transfers angle measurements to the UP board via Serial Peripheral Interface (SPI) with a maximum data transmission rate of almost \SI{20}{\kilo\hertz}.
In this project we used an update rate of \SI{1}{\kilo\hertz}.
\section{Estimation}\label{sec:estimation}
State estimation involves determining the relative roll angle and angular velocity of the dualcopter with respect to the disk.
There are two steps involved in estimating the state: angle measurement and filtering.
In the angle measurement step, events are processed using a sliding-window Hough transform line detection algorithm \cite{houghtransform} to determine the angle of the horizon relative to the dualcopter's frame.
In the second step, a Kalman filter \cite{kalmanfilter} is used to determine the final state estimates based on the measured angle and the previous state.

\begin{figure}[t]
    \centering
    \begin{subfigure}[t]{\columnwidth}
        \centering
        \includegraphics[width=0.85\columnwidth]{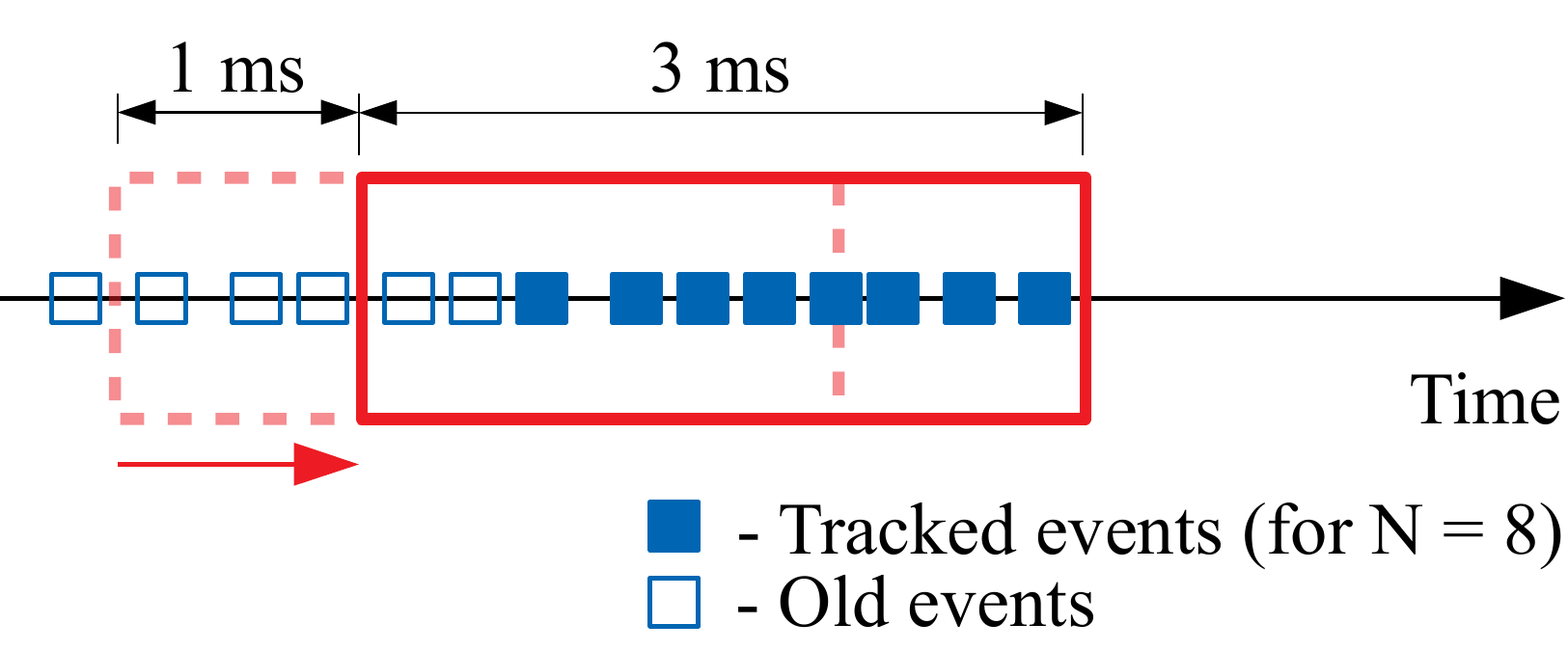}
        \caption{Fast relative motion}
        \label{img:hough_fast}
    \end{subfigure}
    \begin{subfigure}[t]{\columnwidth}
        \centering
        \includegraphics[width=0.85\columnwidth]{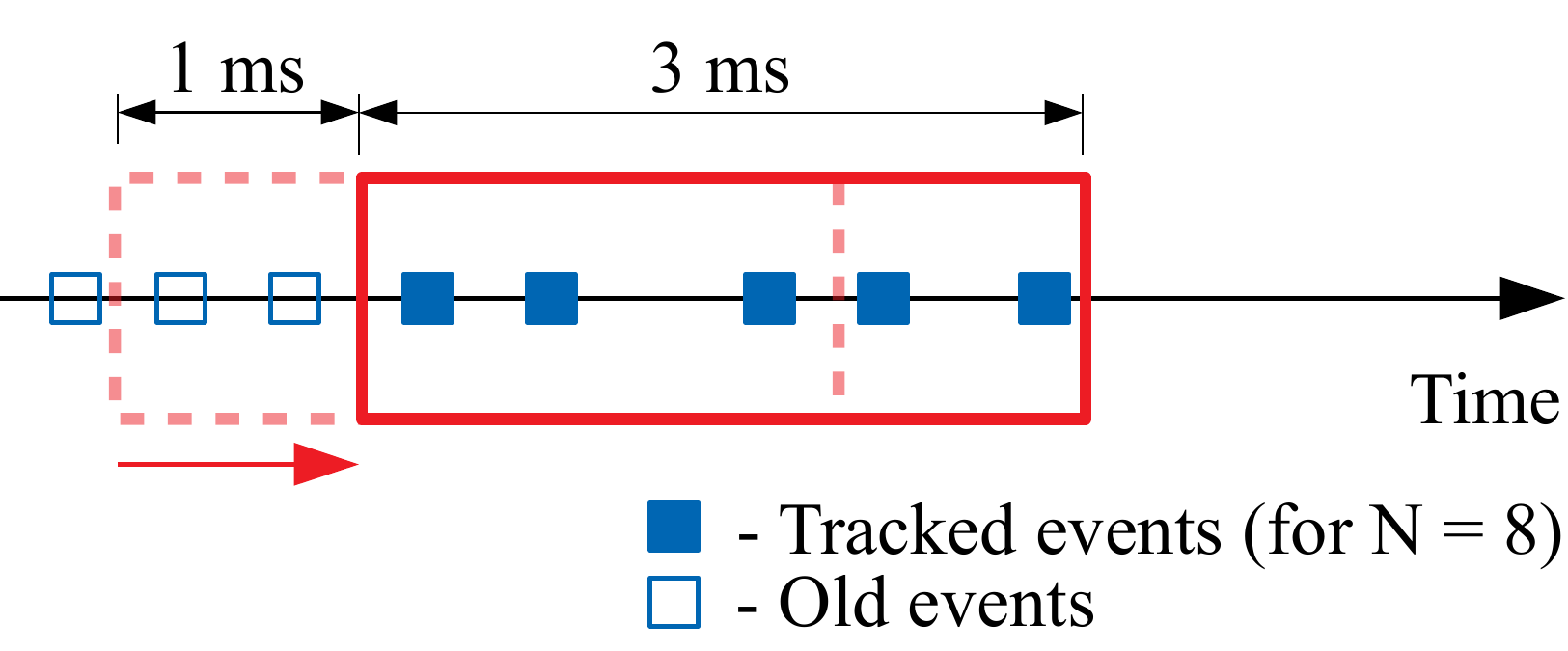}
        \caption{Slow relative motion}
        \label{img:hough_slow}
    \end{subfigure}
     \caption{Sliding-window Hough transform algorithm. The algorithm takes the $N$ most recent events that fit in a \SI{3}{\milli\second} window and uses these to update the state every \SI{1}{\milli\second}. For fast relative motion when many events are generated, a full $N$ events are tracked, while for slower relative motion fewer events are generated and the number of events tracked might be less. Note that the figures above illustrate the case for $N=$ 8 for clarity, while we use $N=$ 80 in our state estimator.}
    \label{img:hough}
\end{figure}

The sliding-window Hough transform line detection algorithm is similar to that implemented in \cite{mueggler2014}, and is illustrated in Fig.~\ref{img:hough}.
The algorithm tracks a constant number of events $N=$ 80 to give an accurate state estimate with little effect from noise.
The window size is capped to \SI{3}{\milli\second} to minimize the accumulation of old events.

The Hough transform algorithm works by parametrizing lines in terms of two parameters, $\rho$ and $\theta$, and iterating through them to find the line that fits the maximum number of events \cite{houghtransform}.
The Hough space accumulator is discretized (in the $\rho$-$\theta$ space) into bins of size \SI{5}{\degree} by 5 pixels.
At a rate of \SI{1}{\kilo\hertz}, i.e., every millisecond, new incoming events are used to update the Hough space and old events are removed in order to maintain a maximum number of $N=$ 80 events.
The attitude of the disk with respect to the dualcopter is considered to be the $\theta$ value of the bin with the maximum count in the updated Hough space.
We require that there are at least 40 events lying along the line detected to prevent false detection due to noise.
The relative angle of the dualcopter with respect to the disk is related to the line parameter $\theta$ by a simple sign change.

The Kalman filter takes the angle estimate from the Hough transform as the measurement to update its two states: angular position and angular velocity.
The system's dynamic model is given by:
\begin{equation}
  \begin{vmatrix}
  \alpha_{k+1}\\
  \dot{\alpha}_{k+1}\\
  \end{vmatrix} = 
  \begin{vmatrix}
  \alpha_k + \dot{\alpha}_k \Delta t\\
  \dot{\alpha}_k + u_k\\
  \end{vmatrix}
  \label{eq:kfmodel}
\end{equation}
where $\alpha_k$ and $\dot{\alpha}_k$ are the relative roll angle and angular velocity at time step $k$, $\Delta t$ is the time interval between time steps $k$ and $k+1$, and $u_k$ is the input angular velocity corresponding to the motor commands sent at time step $k$.

The covariances of the Kalman filter were experimentally tuned to ensure sufficient noise reduction while maintaining fast response to changes in the angular velocity.
The final process and measurement covariances are:
\begin{equation}
  Q = 
  \begin{vmatrix}
  1&0\\
  0&10000\\
  \end{vmatrix}
\hspace{1cm}
R = 10\label{eq:covariances}
\end{equation}
\vspace{0.1cm}
\section{Control}\label{sec:control}
We implemented a simple PD attitude controller for the horizon tracking.
The PD controller takes in the relative angular position and angular velocity estimates from the Kalman filter and calculates the rotor thrusts required for the dualcopter to achieve the desired roll angle.
We simplify the model of the dualcopter by assuming that the center of mass is located at the roll axis.
The equation of motion is then given by:
\begin{equation}
J \ddot{\alpha} = T \label{eq:eom}
\end{equation}
where $J$ is the moment of inertia and $T$ is the torque about the roll axis.

The PD controller determines the torque required $T^*$ by comparing the current relative roll angle $\alpha$ and angular velocity $\dot{\alpha}$ with the desired values $\alpha_{des}$ and $\dot{\alpha}_{des}$ :
\begin{equation}
T^* = k_p (\alpha_{des} - \alpha) + k_d (\dot{\alpha}_{des} - \dot{\alpha}) \label{eq:pd}
\end{equation}

The controller gains $k_p$ and $k_d$ are related to the time constant $\tau$ and damping ratio $\zeta$ of the controller by:
\begin{equation}
k_p = J / \tau^2 
\hspace{1cm}
k_d = 2 \zeta J / \tau \label{eq:gains}
\end{equation}

From the torque $T^*$ we obtain the required thrust difference between the motors by dividing $T^*$ by %
the dualcopter's arm length.

\begin{figure}[t]
    \centering
    \includegraphics[width=0.48\textwidth]{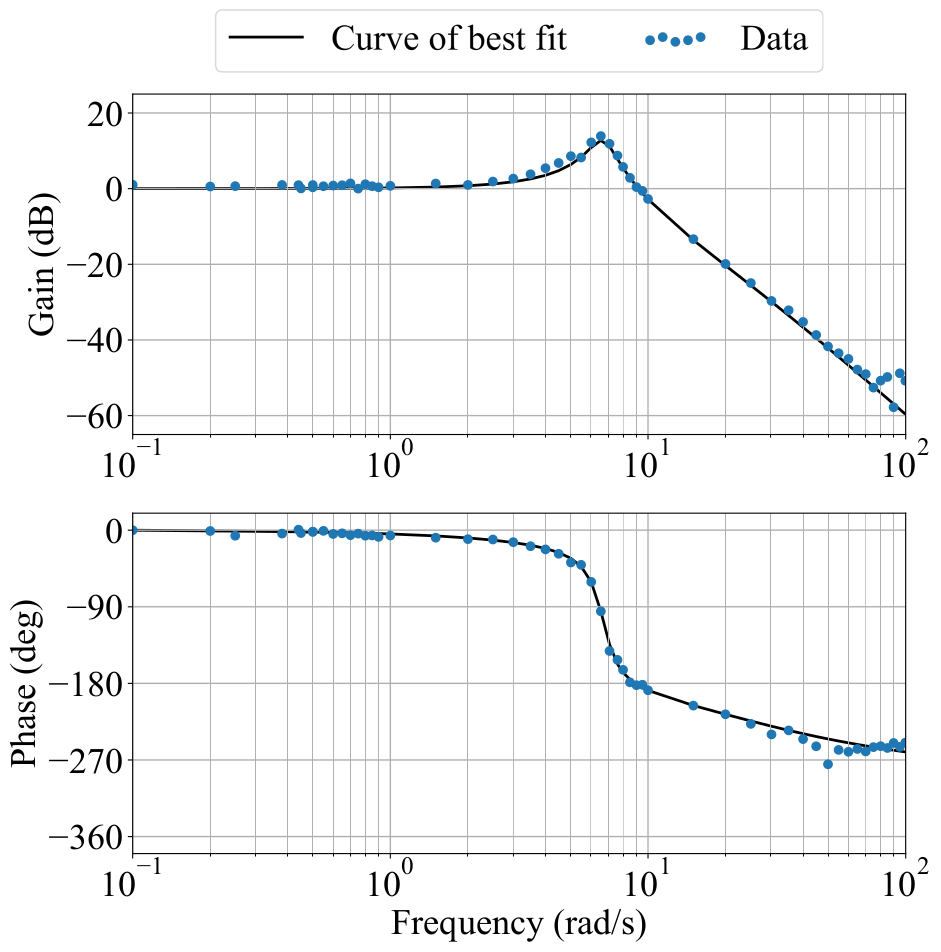}
    \caption{Bode plot of encoder-driven PD controller with gains \hl{$k_p$ = 0.353 and $k_d$ = 0.012}. The data points are shown in blue. The black lines show the best-fit third-order transfer function, a result of the second-order control plant with the effects of first-order motor dynamics observed as well.}
   \label{img:bodeplot_inertia}
\end{figure}

We estimated the moment of inertia of the system through a frequency response test on a PD controller with gains \hl{$k_p$~=~0.353 and $k_d$ = 0.012}, using angle measurements from the encoder on the dualcopter for feedback.
The Bode plot of this controller is shown in Fig. \ref{img:bodeplot_inertia}.
We expect the attitude controller to be a second order system. However, the Bode plot presents the gain and phase characteristics of a third-order system, decreasing by \SI{-60}{\decibel\per\decade} and down to \SI{-270}{\degree} for frequencies beyond \SI{10}{\radian\per\second}.
We suspect that this is due to the dynamics of the motors being used.

By fitting a third-order system function to the Bode plot, we determined the natural frequency of this controller to be \SI{6.69}{\radian\per\second}, which corresponds to a time constant of $\tau$~=~\SI{0.149}{\second}.
From this we estimate the moment of inertia of the system to be \highlight{$J$ = \SI{0.00788}{\kilo\gram\meter\squared}}.

The final controller gains are:
\begin{equation}
\highlight{k_p = 0.353 \hspace{1cm} k_d = 0.071} \label{eq:finalgains}
\end{equation}
which give a controller with a time constant of $\tau$ = \SI{149}{\milli\second} and a damping ratio of $\zeta$ = 0.7.
\section{Experimental Results}\label{sec:experiments}

\subsection{Setup}

The experimental setup is shown in Fig. 1.
The dualcopter platform sits high enough in the air to minimize aerodynamic effects from the ground and surrounding structures.
The reference disk is positioned in front of the dualcopter's event camera, at a distance of \SI{10}{\centi\meter}, to fill the camera's field of view.
The rotary encoders provide ground truth angle measurements of the disk and the dualcopter, which are used to evaluate the performance of the event-based state estimator and controller.

\subsection{Estimation}

We evaluated the performance of the event-based state estimator by analyzing its computational time and accuracy.

The computational time and accuracy of state estimation depends on parameters such as the bin size of the discretized Hough space and the number of events processed.
A smaller bin size in the Hough space will allow for greater precision, but will also increase the processing time and thus affect the real-time accuracy of the state estimation.
Finer discretization of the Hough space will also make it difficult to find a common line through all the events.
In this case, the number of events tracked will have to be increased to ensure that there is a clear maximum in the Hough space corresponding to the line of best fit, which will further increase computational time.

Our choice of a \SI{5}{\degree} by 5 pixels bin size in the Hough space ensures a balance between precision and processing time, which also helps improve the accuracy of the system.
With this level of discretization it is enough to track 80 events at a time to be able to clearly identify the line of best fit.
With these parameters the mean computational time of the state estimation, from the time the Hough transform receives the events to when the Kalman filter produces the state update, is less than \SI{700}{\micro\second}.

To evaluate the accuracy of the state estimation, we placed the dualcopter in front of the disk with the horizon at \SI{0}{\degree}.
As the dualcopter rolls, it estimates its roll angle relative to the stationary horizon, while the rotary encoder on the dualcopter measures the ground truth angles.

\begin{figure}[t]
    \centering
    \includegraphics[width=0.48\textwidth]{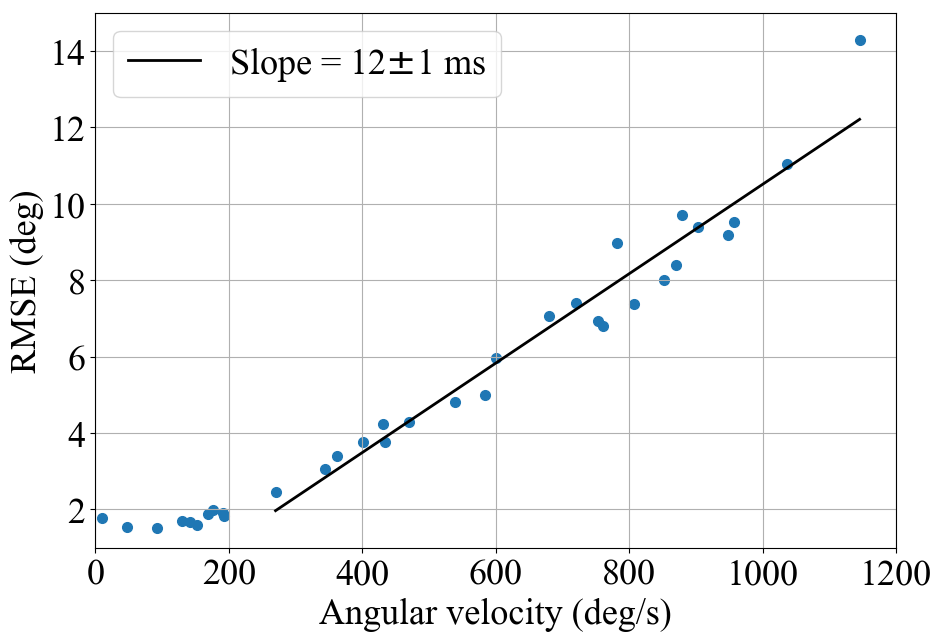}
    \caption{Root mean square error (RMSE) of the angle estimates versus angular velocity of the dualcopter. The regression line (black), fitted through points with RMSE greater than the delay-free RMSE estimate of \SI{2.0}{\degree}, has a slope of \SI{12\pm1}{\milli\second}. The slope gives an estimate on the state-estimation delay.}
   \label{img:rmse_data}
\end{figure}

We observed that the root mean square error (RMSE) of the estimated angles compared to the ground truth angle measurements increases with angular velocity, as shown in Fig.~\ref{img:rmse_data}.
Slow movements of the dualcopter, up to angular velocities of \SI{200}{\degree\per\second}, allow accurate tracking with RMSE less than \SI{2}{\degree}.
This is consistent with the precision of the state estimator, which is approximately \SI{2.5}{\degree} due to the discretization of the Hough space.
Faster movements of the dualcopter result in much larger errors:
when the dualcopter is rotating at an angular velocity of \SI{360}{\degree\per\second}, the RMSE is \SI{3.36}{\degree}; at speeds of over \SI{800}{\degree\per\second} the RMSE of the angle estimates is over \SI{8}{\degree}.
The delay in state estimation is responsible for the increasing error with velocity: if the angle changes rapidly, the actual angle will deviate away from the delayed angle estimate.
Thus the total delay can be determined from the slope of the line relating RMSE to angular velocity.
From the data presented in Fig.~\ref{img:rmse_data} we can deduce that the delay is \SI{12\pm1}{\milli\second}.
However, the expected latency in the vision pipeline is less than this.
The event update interval is \SI{1}{\milli\second}, and the transmission of raw events via USB takes approximately \SI{5}{\milli\second} \cite{usbdelay}.
Since the ground truth angle from the rotary encoder is updated every \SI{1}{\milli\second}, we should expect the event-based angle estimate to be delayed by \SI{5}{\milli\second} from the ground truth measurement.
The discrepancy between the measured and expected state estimation delay suggests that there is  some delay elsewhere in the system that is not being accounted for.
Further investigation is necessary to understand exactly where the delay is coming from.

\subsection{Control}

We evaluated the performance of the vision-driven controller by comparing it to the encoder-driven controller, the best performing controller on the current platform.
The vision-driven controller uses the event-based state estimator to determine the roll angle and angular velocity relative to the horizon marked by the disk, which is stationary here,
while the encoder-driven controller uses the angular measurements from the encoder as feedback to achieve the commanded angle.
The state update rate is \SI{1}{\kilo\hertz} for both controllers, and both use a PD attitude controller with the same gains as in \eqref{eq:finalgains}.

\begin{figure}[t]
    \centering
    \includegraphics[width=0.48\textwidth]{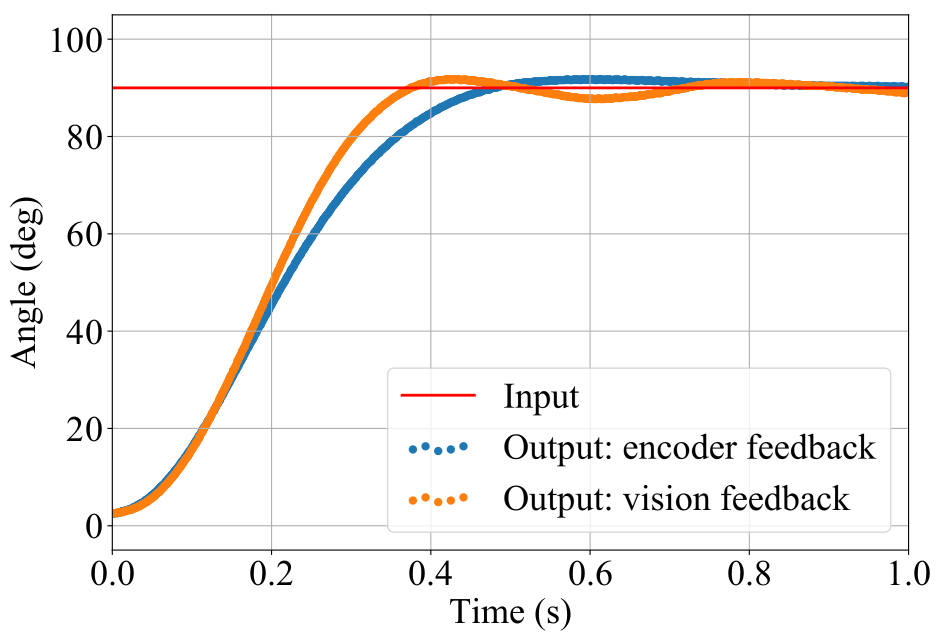}
    \caption{Comparison of the step response (angle = \SI{90}{\degree}) between the vision-driven (orange) and encoder-driven (blue) controllers. The slight differences observed between the two controllers can be explained by the discretizaiton of the angles and latency in the vision pipeline.}
    \label{img:stepresp}
\end{figure}

\begin{figure}[t]
    \centering
    \includegraphics[width=0.48\textwidth]{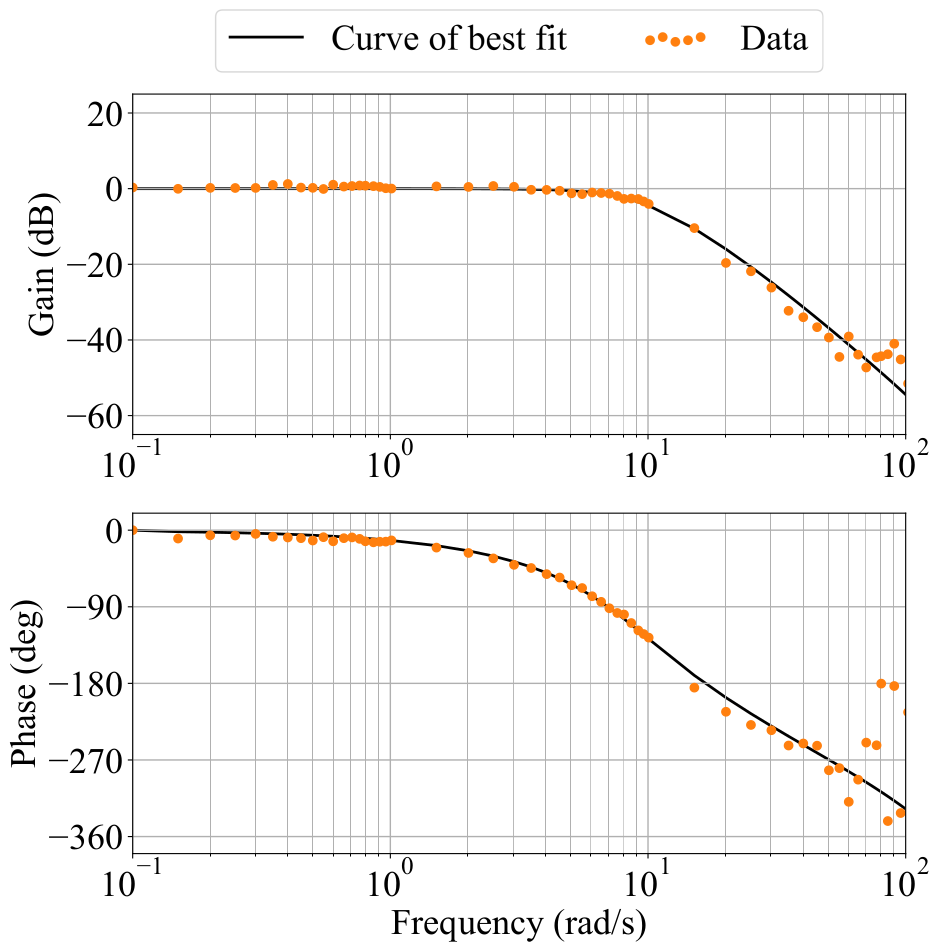}
    \caption{Bode plot showing the frequency response of the final vision-driven controller \hl{($k_p$ = 0.353, $k_d$ = 0.071)}. The data points are shown in orange. The black lines show the best-fit third-order transfer function with a delay determined to be \SI{13.8}{\milli\second}.}
   \label{img:bodeplot_vis}
\end{figure}

\begin{figure}[t]
    \centering
    \includegraphics[width=0.48\textwidth]{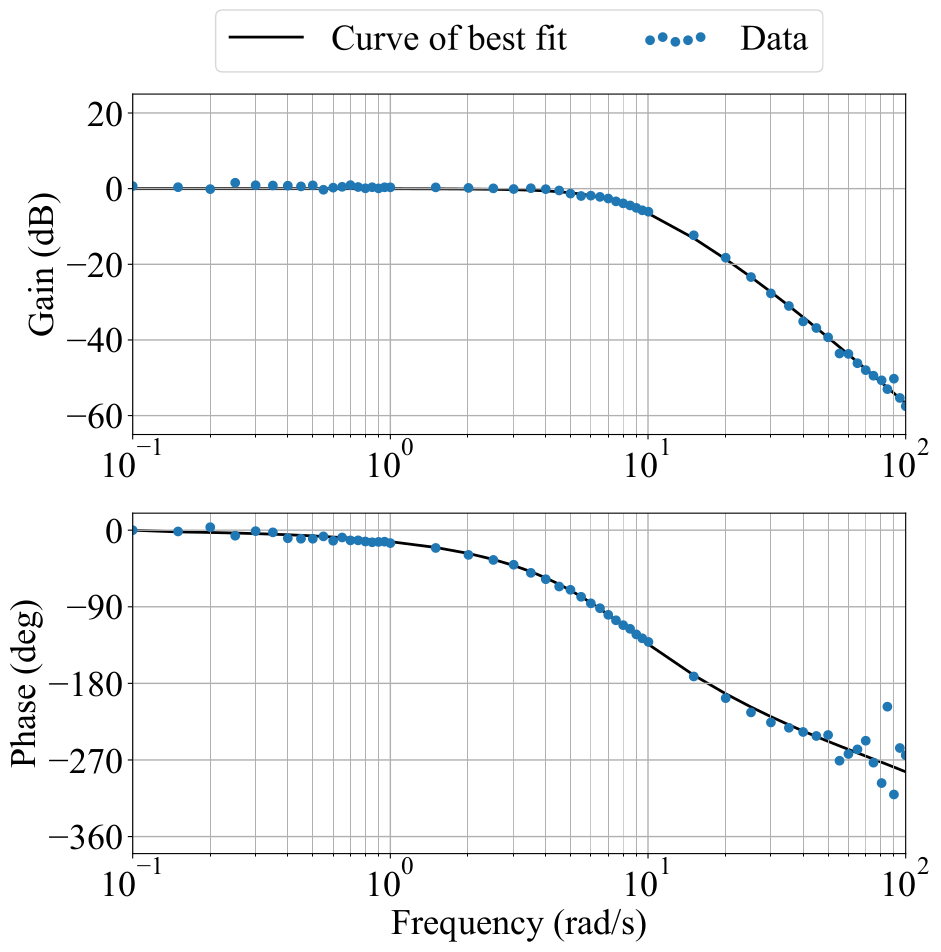}
    \caption{Bode plot showing the frequency response of the final encoder-driven controller \hl{($k_p$ = 0.353, $k_d$ = 0.071)}. The data points are shown in blue. The black lines show the best-fit third-order transfer function with a delay determined to be \SI{5.9}{\milli\second}.}
   \label{img:bodeplot_enc}
\end{figure}

We conducted experiments with various test inputs to the system, including step inputs and sinusoidal inputs of various frequencies.
A plot of the step response is presented in Fig.~\ref{img:stepresp}.
The results from the sinusoidal inputs are summarized in the Bode plots in Fig.~\ref{img:bodeplot_vis} and \ref{img:bodeplot_enc}.

The step response in Fig. \ref{img:stepresp} shows that the output of the vision-driven controller fluctuates when the dualcopter reaches the steady-state roll angle of \SI{90}{\degree}.
When the dualcopter position is held constant and the disk is stationary, the event camera does not generate enough events for the angle estimate to be updated.
Since the dualcopter is not perfectly in balance (the rotors generate slightly different thrust levels, and the center of mass of the dualcopter is not exactly at the center of rotation), there is some drift in the roll angle when the dualcopter receives no feedback.
This effect of reduced number of events coupled with the discretization of the Hough space explains the fluctuations in roll angle observed.

The difference in the rise time of the two plots in Fig.~\ref{img:stepresp} shows the effects of sensor latency.
The measurement delay in the vision pipeline is almost \SI{5}{\milli\second}, while that of the rotary encoder is approximately \SI{1}{\milli\second}.
The larger delay in event transmission results in a delay in event-based state estimation.
This increases the estimated error between the current angle estimate and the commanded angle, which explains the larger angular accelerations of the dualcopter.

The Bode plots in Fig.~\ref{img:bodeplot_vis} and \ref{img:bodeplot_enc} show a summary of the frequency response characteristics of the two controllers.
The control plant is modelled by a third-order system with delay, and we used the Python implementation of the Nelder-Mead simplex algorithm\footnote{https://docs.scipy.org/doc/scipy/reference/generated/\\\indent scipy.optimize.fmin.html} to compute the function parameters that minimize the sum of absolute differences between the data points and the function curve in the Bode plots.
Through the curve fitting we determined the delay in the vision-driven controller to be \SI{13.8}{\milli\second}, and the delay in the encoder-driven controller to be \SI{5.9}{\milli\second}.
Given that the delay for vision-based state estimation is \SI{12}{\milli\second} and the transmission of motor commands from the UP board through the microcontroller and then to the ESC takes less than half a millisecond, the delay for the full vision-driven controller determined from the fit closely matches the expected values.

The results in Fig. \ref{img:stepresp}, \ref{img:bodeplot_vis} and \ref{img:bodeplot_enc} show that overall the response of the vision-driven controller closely resembles that of the encoder-driven controller, despite the latency in the vision pipeline and the error in state estimation.
This indicates that we are reaching the physical limits of the dualcopter platform:
the actuators must be upgraded and the system latency reduced if we want to improve the performance of the controller any further.

A video demonstration of the horizon tracking dualcopter has been submitted together with this work.
In the video we can see that the dualcopter is able to follow the horizon even when the disk is rotated at speeds of over \SI{1600}{\degree\per\second}.

\section{Conclusion and Outlook}\label{sec:conclusion}

In this work we implemented a horizon tracker on a dualcopter platform as a first step towards low-latency high-bandwidth control of drones with event cameras.
The horizon tracking results show that event-based state estimation can be effectively used for closed loop control: even with a simple PD controller the dualcopter is able to track the horizon at high speeds of over \SI{1600}{\degree\per\second}. 

The current state estimator, with an update rate of \SI{1}{\kilo\hertz}, exceeds the bandwidth of the control plant.
The platform can be further improved for speed and accuracy by reducing the delay in the system.
For instance, sensor latency can be reduced by replacing the DAVIS with an embedded Dynamic Vision Sensor (eDVS).
The eDVS has an embedded computer that can process the events onboard and send the results via UART, which is much faster than the transmission of raw events via USB.
Upgrading the actuators will also help increase the maximum possible acceleration of the dualcopter to enable even more aggressive control.

In the future, we intend to use the dualcopter platform to explore event-based control, in which events directly encode control signals, thereby removing the state estimation step altogether.
Such algorithms have been investigated in \cite{singh2016} and \cite{singh2019}, and they have yet to be tested on a physical platform.
The experimental setup developed in this work provides a platform to which various event-based control algorithms can be adapted for testing.

We also plan to implement event-camera-based closed loop control on a hovering quadcopter, taking the project one step closer to our ultimate goal of fast, agile flight solely based on event-camera feedback.

\section*{ACKNOWLEDGMENT}\label{sec:acknowledgment}
We gratefully acknowledge Elias Mueggler and Davide Falanga for building the original dualcopter platform as well as Thomas Laengle and Balazs Nagy for support with the hardware setup.

\addtolength{\textheight}{-12cm}   %

\IEEEtriggeratref{9}
\bibliographystyle{ieeetr}

\end{document}